\documentclass[conference]{IEEEtran}

\usepackage{spconf,amsmath,graphicx}
\usepackage{color}
\usepackage{cite}
\usepackage{graphicx}
\usepackage{hyperref}
\hypersetup{
     colorlinks   = true,
     urlcolor = blue
}

\makeatletter

\def\ps@IEEEtitlepagestyle{
  \def\@oddfoot{\mycopyrightnotice}
  \def\@evenfoot{}
}
\def\mycopyrightnotice{
  {\footnotesize } 
  \gdef\mycopyrightnotice{}
}

\makeatother

\usepackage{eso-pic}
\newcommand\AtPageUpperMyright[1]{\AtPageUpperLeft{
 \put(\LenToUnit{2cm},\LenToUnit{-1cm}){
     \parbox{1\textwidth}{\raggedright\fontsize{8}{11}\selectfont #1}}
 }}
\newcommand{\conf}[1]{
\AddToShipoutPictureBG*{
\AtPageUpperMyright{#1}
}
}

\title{Low bit-rate speech coding with VQ-VAE and a WaveNet decoder}

\conf{ICASSP 2019-2019 IEEE International Conference on Acoustics, Speech and Signal Processing (ICASSP), pp. 735-739. IEEE, 2019. DOI: \href{https://doi.org/10.1109/ICASSP.2019.8683277}{10.1109/ICASSP.2019.8683277}. \copyright 2019 IEEE. Personal use of this material is permitted. Permission from IEEE must be obtained for all other uses, in any current or future media, including reprinting/republishing this material for advertising or promotional purposes, creating new collective works, for resale or redistribution to servers or lists, or reuse of any copyrighted component of this work in other works.} 
\name{\begin{tabular}{c}Cristina G\^arbacea\sthanks{Work performed while at DeepMind}$^{1}$, A{\"a}ron van den Oord$^{2}$, Yazhe Li$^{2}$,\\Felicia S C Lim$^{3}$, Alejandro Luebs$^{3}$, Oriol Vinyals$^{2}$, Thomas C Walters$^{2}$\end{tabular}}

\address{$^{1}$University of Michigan, Ann Arbor, USA $^{2}$DeepMind, London, UK $^{3}$Google, San Francisco, USA} 
\begin{document}
\thispagestyle{IEEEtitlepagestyle}
%

\maketitle
\IEEEpubidadjcol
\begin{abstract}

In order to efficiently transmit and store speech signals, speech codecs create a minimally redundant representation of the input signal which is then decoded at the receiver with the best possible perceptual quality. In this work we demonstrate that a neural network architecture based on VQ-VAE with a WaveNet decoder can be used to perform very low bit-rate speech coding with high reconstruction quality. A prosody-transparent and speaker-independent model trained on the LibriSpeech corpus coding audio at 1.6 kbps exhibits perceptual quality which is around halfway between the MELP codec at 2.4 kbps and AMR-WB codec at 23.05 kbps. In addition, when training on high-quality recorded speech with the test speaker included in the training set, a model coding speech at 1.6 kbps produces output of similar perceptual quality to that generated by AMR-WB at 23.05 kbps.
\end{abstract}

\begin{keywords}
Speech coding, low bit-rate, generative models, WaveNet, VQ-VAE 
\end{keywords}
\section{Introduction and Related Work}
\label{sec:intro}

Speech codecs typically employ a carefully hand-engineered pipeline made up of an encoder and a decoder which take into account the physics of speech production to remove redundancies in the data and yield a compact bitstream. High quality speech codecs typically operate at bit-rates over 16 kbps. Recent advances in the field of deep learning, specifically autoencoder networks which directly learn the mapping of inputs to outputs by means of an encoder-decoder framework with an information bottleneck in between, open up the possibility of learning both the encoder and the decoder directly from speech data. These models are able to learn the redundancies in signals directly by being exposed to many examples during training, and have been successfully applied in the domain of image compression \cite{theis2017lossy, DBLP:journals/corr/BalleLS16a, DBLP:journals/corr/AgustssonMTCTBG17,rippel2017real}. In the speech domain, end-to-end coding operating at rates as low as 369 bps by use of extracted acoustic features \cite{cernak2016composition} and an end-to-end optimized wideband codec performing on par with the adaptive multirate wideband (AMR-WB) codec at 9-24 kbps \cite{kankanahalli2018end} have both recently been proposed.
Creating a minimally redundant representation of speech is inherently related to determining the true information rate of the input signal. Kuyk \emph{et al.} \cite{van2017information} compute the true information rate of speech to be less than 100 bps, yet current systems typically require a rate roughly two orders of magnitude higher than this to produce good quality speech, suggesting that there is significant room for improvement in speech coding.

The WaveNet \cite{van2016wavenet} text-to-speech model shows the power of learning from raw data to generate speech. Kleijn \emph{et al.} \cite{chrome} use a learned WaveNet decoder to produce audio comparable in quality to that produced by the AMR-WB \cite{bessette2002adaptive} codec at 23.05 kbps from the bit stream generated by the encoder in Codec2 \cite{rowe2011codec} (a parametric codec designed for low bit-rates) operating at 2.4 kbps; this demonstrates the effectiveness of a learned decoder over a hand-engineered one.
Furthermore, van den Oord \emph{et al.} \cite{vqvae} demonstrate a learned autoencoder -- the vector-quantized variational autoencoder (VQ-VAE) -- which is able to encode speech into a compact discrete latent representation and then reconstruct the original audio with high quality by sampling from a WaveNet-like decoder.

In this paper, we evaluate the VQ-VAE architecture introduced in \cite{vqvae} as an end-to-end learned speech codec, showing that it can yield high reconstruction quality while passing speech through a compact latent representation corresponding to very low bit-rates for coding. We propose and evaluate various modifications to the VQ-VAE / WaveNet architecture as described in \cite{vqvae} in order to make it more suited to the task of speech coding. We create a speaker-independent model which can accurately reproduce both the content and the prosody of the input utterance while passing through a compact latent representation that leads to lower bit rates and higher reconstruction quality than the current state-of-the-art.

The VQ-VAE combines a variational autoencoder (VAE) \cite{kingma2013auto} with a vector quantization (VQ) layer to produce a discrete latent representation which has been shown to capture important high-level features in image, audio and video data, yielding an extremely compact and semantically meaningful representation of the input. The prior and posterior distributions are categorical, and samples drawn from these distributions index an embedding which is passed as input to the decoder network. 
Van den Oord \emph{et al}. \cite{vqvae} demonstrate the use of VQ-VAE in the audio domain, using a convolutional encoder and a WaveNet decoder on speech. The authors use the VQ-VAE model to generate very high-quality speech even when using a latent representation that is 64 times smaller than the original input waveform, reducing 16-bit pulse-code modulation (PCM) encoded speech at 16 kHz sample rate (256 kbps) to a stream of discrete latent codes at 4 kbps.
In this case, the network learns to represent the high-level semantic content of the speech, therefore when a speech waveform is passed through the autoencoder network, the reconstruction contains the same syllabic content, but the prosody of the utterance may be significantly altered. Further analysis of the representation learned by the network shows that the discrete latent codes are well-correlated with the stream of syllables in the input speech, suggesting that the encoder network builds a high-level representation of the content of the utterance, and then the decoder generates plausible-sounding specific details. Given the very high quality of the reconstructed audio, and the compact latent representation that the network can produce, the VQ-VAE architecture seems to lend itself naturally to the task of low bit-rate audio coding if it can be generalized to preserve the identity of arbitrary speakers.
\section{VQ-VAE BASED Speech Codec}
\label{sec:model_architecture}

Given the ability of the VQ-VAE model to learn a high-level abstract space invariant to low-level perturbations in the input and which encodes only the contents of speech \cite{vqvae}, our main goal is to determine if the model can be used as an end-to-end speech codec (i.e. map the input audio to the discrete latent space learnt by VQ-VAE and then reconstruct the signal by sampling from the decoder of the network conditioned on the latent representation). In order to be useful as a generic speech codec, the model must be constrained to maintain both the speaker identity and the prosody of the utterance at decoding time. We present these architectural changes below.

\subsection{Maintaining speaker identity}
\label{sec:speaker_identity}


To encourage the model to be speaker-agnostic and generalize across speakers it has never seen in the training set, we modify the original VQ-VAE architecture by removing explicit conditioning on speaker identity, which was done through a one-hot code passed to both the encoder and the decoder both at training and synthesis time. In its place, we add a latent representation (with an associated codebook) that takes its input from the whole utterance and does not vary over time. The time invariant code is generated by mean pooling over the time dimension of the encoder output and fed to a separate codebook. The expectation is that the network will learn to use the time-varying set of codes to encode the message content which varies over time, while summarising and passing speaker-related information through the separate non time-varying set of codes. In our experiments, the non time-varying code is computed over the entire utterance at encoding time. In the experiments below, the codebooks were of similar size to those used by the time-varying latents (eg. 256 elements), which led to a small additional coding overhead of 1-2 bytes over the course of the utterance. Such a long window would be unrealistic for online speech coding; in this case the static code could be replaced with a code that varies slowly over time and which is based only on past information.
\subsection{Constraining prosody}
\label{sec:constraining_prosody}
 In order to constrain the model to pass prosodic as well as semantic information through the bottleneck representation, it is necessary to design a training regime in which the network is encouraged, via a loss term, to pass pitch ($f_{0}$) and timing information through to the decoder. To this end, a second decoder was added to the network in parallel with the WaveNet audio decoder; the task of this network is to predict the $f_{0}$ track of the training utterance, using a common latent representation with the audio decoder. An additional $f_{0}$ prediction term with a tuneable weight is added to the loss function, which causes the latent representation to pass pitch track information through the bottleneck. This information is then available to the waveform decoder network, which uses it to produce an utterance that has the same pitch track as the original. The $f_{0}$ predictor model uses the same architecture as the audio decoder, but only has to upsample the latent representation to the rate of the $f_{0}$ feature. When the trained model is used as a codec, the $f_0$ prediction network can be removed from the model, and the audio decoder alone used to reconstruct the waveform.

In our datasets $f_0$ information had already been extracted from the audio waveform using a standard pitch tracker, and was sampled at 200 Hz. Given the 16 kHz sample rate of the audio in the training set, we introduce an extra convolutional layer with stride 5 on the encoder side for rate compatibility between the audio samples and the $f_0$ samples.
\section{Experiments}
\label{sec:experiments}

In the set of experiments below our goal is to determine under which hyper-parameter settings VQ-VAE can be used as a very low rate speech codec. Therefore, we focus on establishing how the quality of the reconstructed signal varies as a function of the latent representation size of the model. Changing the dimensionality of the latent representation is equivalent to varying the bit-rate of the encoded signal. In the original VQ-VAE setup the latents consist of one feature map and the latent space is 512-dimensional. In our experiments we vary in turn the number of encoder layers, the number of latent maps, as well as the number of discrete codes per map.

\textbf{Datasets}
We carry our experiments on two different datasets. Most experiments rely on the LibriSpeech \cite{panayotov2015LibriSpeech} corpus (LS), which contains around 1,000 hours of read English speech from a total of 2,302 speakers at a sample rate of 16 kHz. For the majority of results presented in this paper, the \texttt{train-clean-100}, \texttt{train-clean-360} and \texttt{train-other-500} subsets of the corpus were combined and used at training time. Evaluation samples were created using the \texttt{test-clean} set. The set of speakers in the test data is disjoint with those in the training data.

For cases when we evaluate models including test speakers at training time, we held out 1/100 of the \texttt{test-clean} dataset for evaluation, and add the remaining data (totalling $\sim$5 hours of extra audio) to the training set. We refer to this dataset as the augmented LibriSpeech (LSplus) corpus. The datasets were annotated with automatically extracted $f_0$ information (presented as $\mathrm{log}(f_0)$), sampled at 200 Hz.

In addition to the datasets presented above, we also had access to a proprietary corpus consisting of 10 American English speakers totalling approximately 300 hours of speech, of which at least 85\% was recorded in studio conditions. This dataset, referred to as the Studio corpus, is used to investigate performance in ideal training conditions.

\textbf{Evaluation} To assess the reconstructed speech samples in a way that reflects informal listening impressions, we follow Kleijn \emph{et al.} \cite{chrome} and use subjective MUSHRA-type listening tests \cite{recommendation2001method} to measure the perceived quality of the output from lossy audio compression algorithms. In MUSHRA evaluations the listener is presented with a labelled uncompressed signal for reference, a set of numbered samples including the test samples, a copy of the uncompressed reference, and a low-quality anchor. We compress the anchor samples with Speex \cite{valin2016speex} at 2.4kbps (below its comfortable range of use).

For some evaluations we also include other speech codecs for comparison: Codec2 \cite{rowe2011codec} at 2.4kbps, MELP \cite{mccree19962} at 2.4kbps, AMR-WB \cite{bessette2002adaptive} at 23.05kbps and the original signal after companding and quantization using 8-bit $\mu$-law \cite{mulaw1988} (giving a bit-rate of 128kbps). $\mu$-law is the companding function used by the WaveNet decoder to map the 16-bit sample resolution to an 8-bit softmax distribution, and so it gives an upper bound on the quality of the reconstruction from the VQ-VAE model. Each evaluation includes 8 utterances, and each utterance is evaluated by 100 human raters. The MUSHRA scale from 0-100 allows rating very small differences between the samples.

\subsection{Quality Evaluation}
\begin{figure}[t]
  \includegraphics[scale=.75,width=\columnwidth]{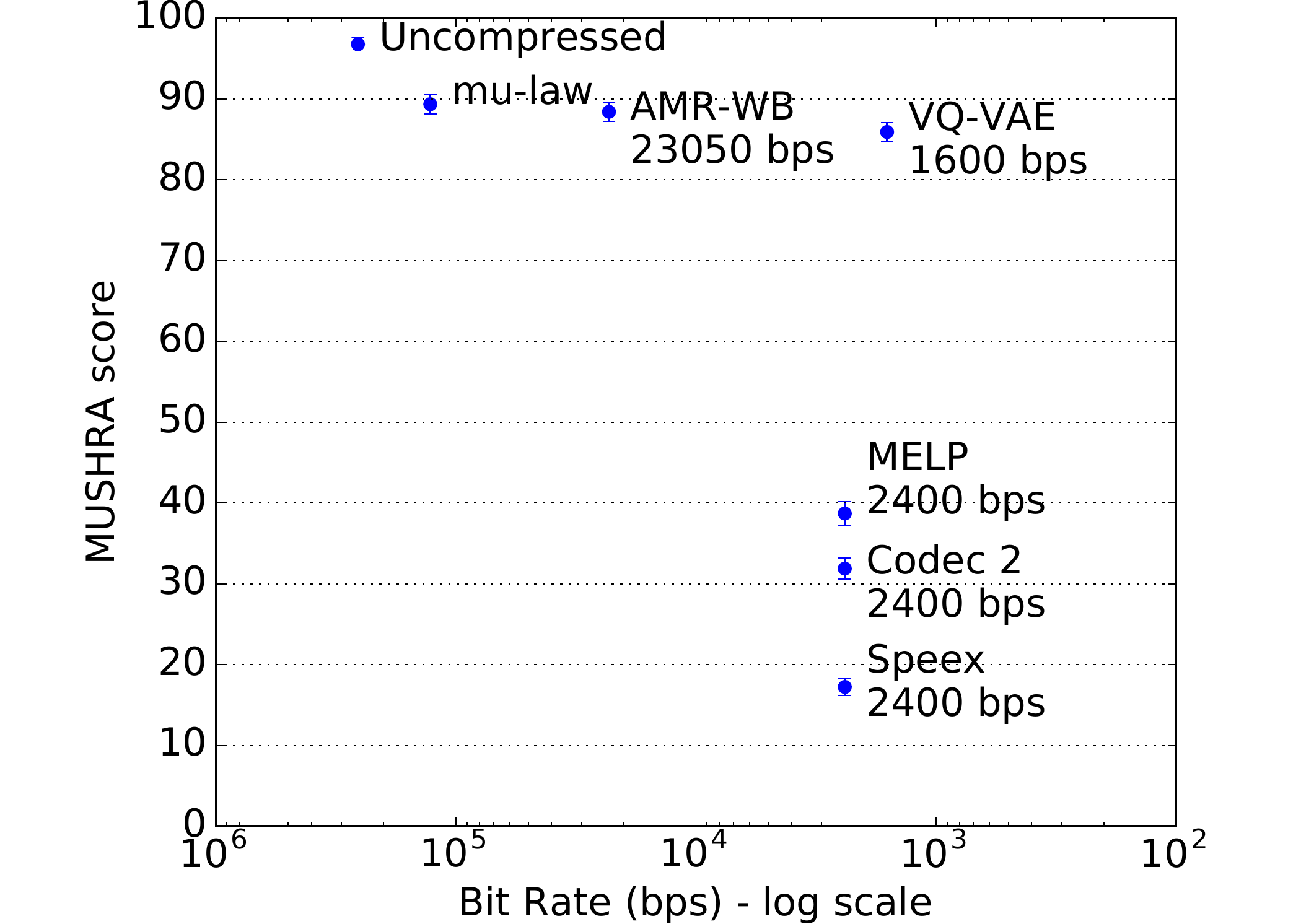}
  \caption{MUSHRA score vs bit-rate for the VQ-VAE speech codec at 1.6 kbps, trained on Studio data and evaluated on a single studio-recorded voice present in the train set, against a variety of other codecs. Bit rate reduces from left to right, with the optimum performance for a codec suggested in \cite{van2017information} being in the top right corner of the graph.}
  \label{mushra_eval_studio}
\end{figure}

Initial experiments with the Studio dataset led to a choice of hyperparameters that gave very good reconstruction quality for voices in that dataset. The number of encoder layers (each layer downsampling by a factor equal to the stride), the number of latent maps (i.e. the number of separate codebooks) and the size of each codebook used were varied. We ascertain that high-quality reconstruction of audio is possible with five strided convolutional encoder layers of stride 2 and one of stride 5 (leading to a total downsampling factor of $2^5 * 5 = 160$ - a feature rate of 100 Hz), two latent maps each with a 256-element codebook (each map is representable by an 8-bit value), coding a 64-dimensional latent vector. This leads to a bit rate of 1600 bps. This model is evaluated in a MUSHRA test against a range of other codecs, and the results are presented in Figure \ref{mushra_eval_studio}.

In the plot we notice the emergence of two clusters of high and low reconstruction quality: \textit{i) the low quality cluster} includes Speex, Codec2 and MELP operating at 2.4kbps, and \textit{ii) the high quality cluster} consists of AMR-WB at 23.05kbps, 8 bit $\mu$-law and VQ-VAE at 1.6kbps. In these conditions, using speech from a speaker contained in the training set and trained on studio-quality utterances, the VQ-VAE model is very effective at coding the audio, achieving a factor of 14 improvement in compression rate over a standard wideband audio codec for a minimal reduction in perceptual quality.

While this is an interesting result, it does not make for a fully robust codec system, since the test speaker was present in the training set, and the train and test audio were both of very high quality. In order to evaluate the performance of the model in more natural conditions, we then switched to using the LibriSpeech corpus instead of the Studio corpus, which also has the benefit of containing many more speakers and a variety of different noise conditions.

\begin{figure}[t]
  \includegraphics[width=\columnwidth]{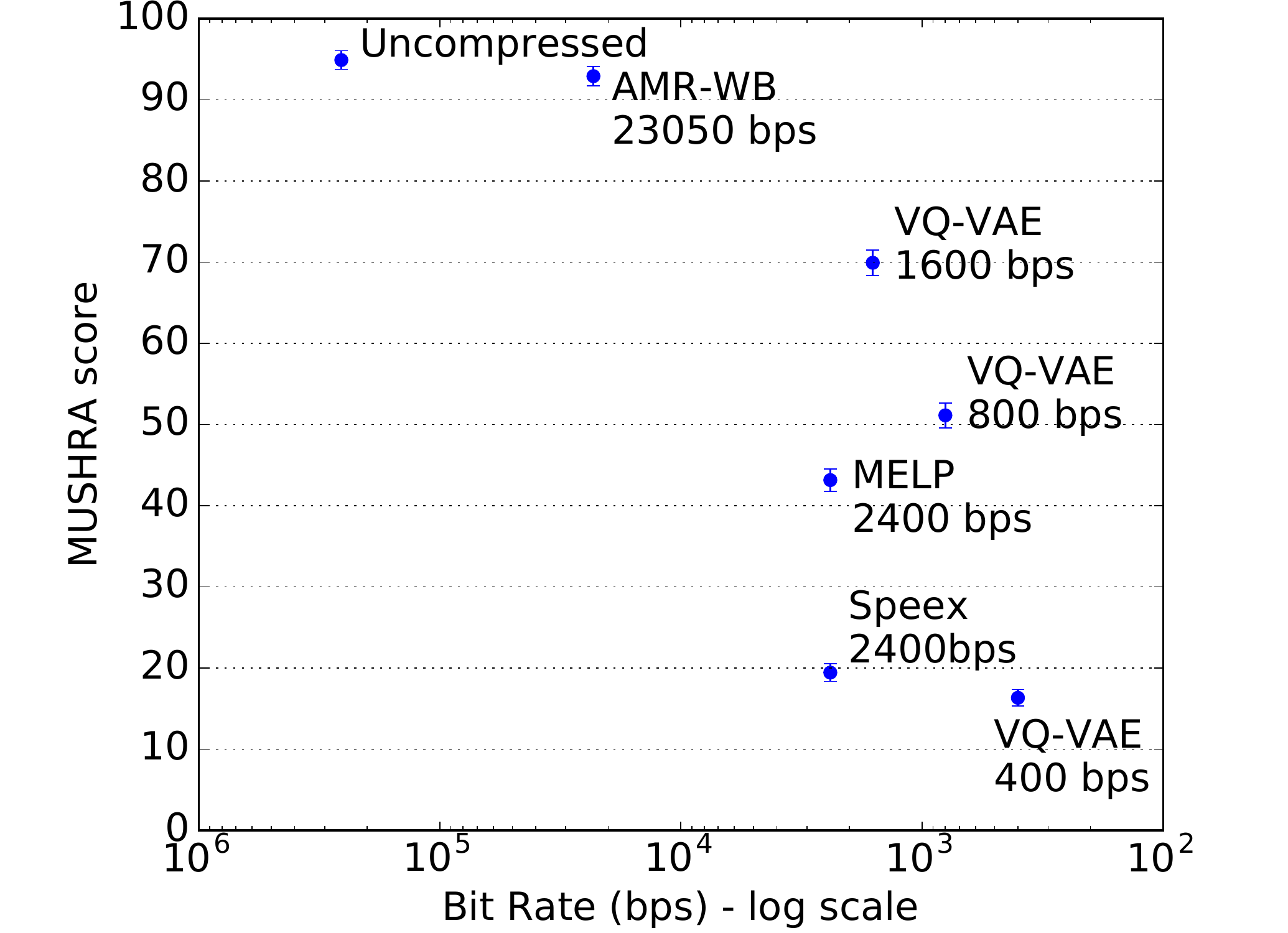}
  \caption{MUSHRA score vs bit-rate for the VQ-VAE speech codec, trained on LibriSpeech data and evaluated on voices from the LibriSpeech test set, against various other codecs.}
  \label{mushra_eval_rate_vs_quality}
\end{figure}

In Figure \ref{mushra_eval_rate_vs_quality} we present results for a set of models trained on the full LibriSpeech dataset. The 1600 bps model uses the same architecture parameters as the model trained on Studio data above, while the 800 bps and 400 bps models each add one further encoder layer with a stride of 2, which halves the bit rate of the model. Here we see that the model at 1600 bps is lower in perceptual quality than previously, but still roughly halfway between MELP at 2400 bps and AMR-WB at 23.05 kbps. Even at 800 bps, the VQ-VAE model outperforms MELP at 3 times the bitrate, and it is only at 400 bps that performance degrades below that of Speex at 2400 bps.

\begin{figure}[t]
  \includegraphics[width=\columnwidth]{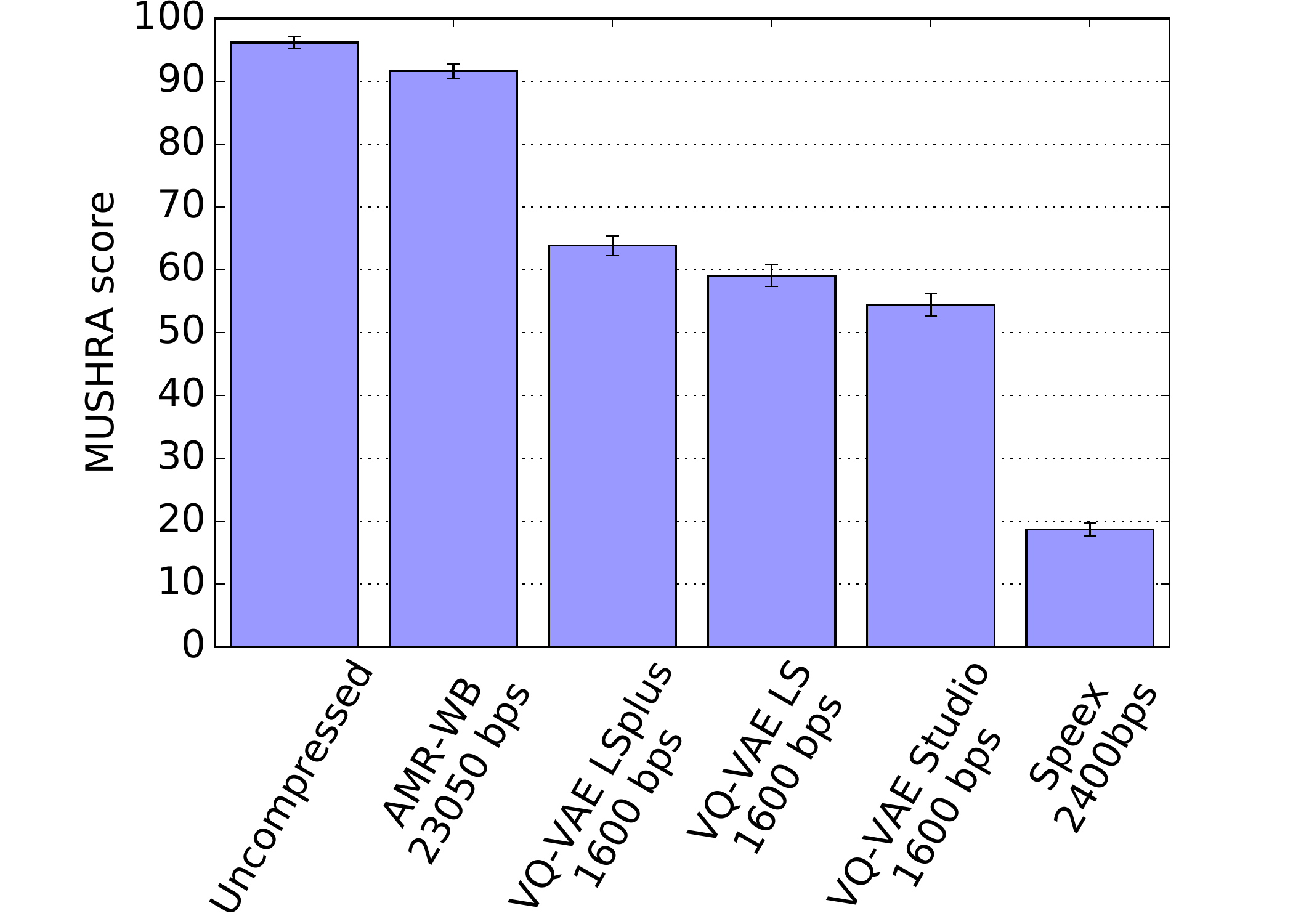}
  \caption{MUSHRA results for the VQ-VAE speech codec at 1.6 kbps, trained on Studio data, on LibriSpeech and on LibriSpeech plus test set speakers and evaluated on voices from the LibriSpeech test set.}
  \label{lsx_ls_studio_mushra}
\end{figure}

Finally, we investigate the effect of including utterances from the test speaker in the train set, to determine how much benefit the model can gain from training on a specific speaker. These results are presented in Figure \ref{lsx_ls_studio_mushra}, in which we observe a small advantage in the perceptual quality for the VQ-VAE model when training on utterances from test speakers.

\subsection{Speaker Transparency}


We follow Jia \emph{et al.} \cite{jia2018transfer} and Chen \emph{et al.} \cite{chen2018sample} who assess speaker similarity using mean opinion score (MOS) evaluations based on subjective listening tests. MOS evaluations \cite{rec1996p} map ratings from bad to excellent into the range 1-5 with 0.5 point increments. To measure how similar synthesized speech is to real speech of the target speaker, each synthesized utterance is paired with a randomly selected ground truth utterance from the same speaker. Raters are explicitly asked throughout the study to not judge the content, grammar or the audio quality of the sentences, but instead to focus on the similarity of speakers to one another. Eight pairs of utterances from the LibriSpeech test set were rated by 20 listeners each.

MOS scores for speaker similarity are presented in Table \ref{mos_eval1} for versions of the VQ-VAE codec at 1.6 kbps trained on the LibriSpeech dataset and the augmented LibriSpeech dataset (LSplus, meaning that the test speakers were in the train set), and for comparison MELP at 2.4 kbps and Speex at 2.4 kbps. 

\begin{table}[ht]
\centering
\begin{tabular}{r|l}
 \textbf{Codec} & \textbf{Speaker Similarity MOS} \\ \hline
 VQ-VAE LSplus 1600 bps &  $3.794 \pm 0.451$ \\
 VQ-VAE LS 1600 bps & $3.703 \pm 0.716$\\
 MELP 2400 bps &  $3.138 \pm 0.324$ \\
 Speex 2400 bps &  $2.534 \pm 0.233$ \\ \hline
\end{tabular}
\caption{MOS results for the VQ-VAE speech codec at 1600 bps trained on LibriSpeech with and without the test voices in the train set, Speex at 2400 bps and MELP at 2400 bps.}
\label{mos_eval1}
\end{table}

We see that the speaker similarity MOS for the VQ-VAE based codec is higher than for the other two low-bitrate codecs, and is slightly higher for the case where the test speakers are included in the train set. The wider error bounds on the result for the VQ-VAE codec trained on LibriSpeech alone are at least in part due to a very low MOS, of 1.85, being assigned to one of the speakers in the evaluation (the next lowest scoring utterance has a speaker similarity MOS of 3.38); the MOS for this speaker is significantly higher, at 2.73, when the test speaker is included in the train set.

\section{Conclusion}
\label{sec:conclusion}

In this work we demonstrate initial results showing the suitability of the VQ-VAE model as a robust very low bit-rate speech codec. After adjusting the architecture to make it speaker and prosody transparent, we show that VQ-VAE at 1600 bps outperforms some popular low bit-rate speech codecs operating at 2400 bps. Listening test results demonstrate that the VQ-VAE based codec is somewhat better at preserving speaker identity than other low-rate codecs, but more work is required to understand the variability in preservation of speaker identity. Additionally we expect that model performance can be improved beyond the current 1600 bps.

\textbf{Acknowledgements} Thanks to Jan Chorowski, Norman Casagrande, Yutian Chen, Ray Smith, Florian Stimberg, Ron Weiss, Heiga Zen and the anonymous ICASSP reviewers.




\bibliographystyle{IEEEbib}
\bibliography{main}

\end{document}